\documentclass{article}

\usepackage{arxiv}
\usepackage[utf8]{inputenc} 
\usepackage[T1]{fontenc}    
\usepackage{hyperref}       
\usepackage{url}            
\usepackage{booktabs}       
\usepackage{amsfonts}       
\usepackage{nicefrac}       
\usepackage{microtype}      
\usepackage{lipsum}
\usepackage{graphicx}
\usepackage{natbib}
\usepackage{xspace}
\usepackage{xcolor}
\usepackage{amsmath}
\usepackage{amssymb}
\usepackage{bm}
\usepackage{placeins}
\usepackage{tabularx}
\usepackage{subcaption}
\usepackage{CJK}
\usepackage[whole]{bxcjkjatype}
\usepackage[utf8]{inputenc}

\newcommand{\jptext}[1]{%
  \begin{CJK}{UTF8}{min}#1\end{CJK}%
}

\usepackage{fancyhdr}
\newcommand{\dimension}[0]{D}
\newcommand{\Dim}[0]{\dimension}  

\newcommand{\threeDots}[0]{...}

\newcommand{\mmdCvAgg}[0]{\text{MMD-CV-AGG}(E_t, E_{t'})}

\newcommand{\setSelectedSenseChangeVariables}[0]{\{S_{t, t'} | t, t' \in \{1,..,T\}, t \neq t' \}}

\newcommand{\bx}{{\bm x}}
\newcommand{\by}{{\bm y}}

\title{Word Sense Detection Leveraging Maximum Mean Discrepancy}
\author{
  Kensuke Mitsuzawa \\
  Université Côte d'Azur, CNRS, LJAD, France \\
  \texttt{kensuke.mitsuzawa@unice.fr} \\
}
\date{\vspace{-5ex}}

\begin{document}
\maketitle
\begin{abstract}
Word sense analysis is an essential analysis work for interpreting the linguistic and social backgrounds.
The word sense change detection is a task of identifying and interpreting shifts in word meanings over time. 
This paper proposes \texttt{MMD-Sense-Analysis}, a novel approach that leverages Maximum Mean Discrepancy (MMD) to select semantically meaningful variables and quantify changes across time periods. 
This method enables both the identification of words undergoing sense shifts and the explanation of their evolution over multiple historical periods. 
To my knowledge, this is the first application of MMD to word sense change detection.
Empirical assessment results demonstrate the effectiveness of the proposed approach.
\end{abstract}


\section{Introduction}
\label{sec:interpretable_two_sample_testing_application_word_embeddings}

Natural language has been developed and changed over time, and, as time goes on, the sense of vocabulary has often changed.
For example, the word \textit{gay} had the sense of \textit{carefree} and has shifted to \textit{homosexual} during the 20th century~\citep{kutuzov-2018}.
In the Natural Language Processing (NLP) community, the task of word sense change analysis detects these changes in vocabulary senses over time.
The task is also called ``semantic change'', ``semantic shift'', ``temporal embeddings'', ``diachronic embeddings'', or ``lexical semantic change''~\cite{Periti-2024-lexical-semantic-change}, however, in this paper, I call it a ``sense change detection'' task, following the terminology~\citep{kutuzov-2018}.

The application of the sense change detection is, for instance, linguistic and social background analysis.
By examining vocabulary that changes over time, we can gain insights into the social backgrounds of the speakers who use those words. 
For instance, the study~\cite{timmermans-2022} investigates sense change detection in a collection of Dutch texts from the 18th and 19th centuries, and infers the historical social context in which nationalism emerged in the Netherlands.

The two major goals of the sense change detection are identifying words that experience the meaning shift between two time periods and explaining the sense change of words over multiple periods~\cite{kiyama-2025}.
To achieve these goals, recent work employs vector representations trained by word embedding models~\cite{Aida-2021} or large language models~\cite{Lazaridou-2021, Loureiro-etal-2022-timelms} for obtaining representations capturing wide contexts of words.
On trained vector representations, there can be two types of dimensions (variables)~\cite{Aida-bollegala-2024-semantic}; \textit{sense-aware dimensions} that are involved with the sense changes and \textit{uniform dimensions} that are not involved.
Since conventional distance metrics (i.e. cosine or euclidean distances) are not able to distinguish these two types of dimensions, \cite{Aida-bollegala-2024-semantic} addresses the importance of a distance metric that can accurately measure the semantic change.

This paper proposes \texttt{MMD-Sense-Analysis} for realizing the two goals of identifying word sense changes and explaining the sense changes over periods, while distinguishing the sense-aware dimensions.
The core idea of \texttt{MMD-Sense-Analysis} is selecting semantic-aware dimensions using a variable selection based on \textit{Maximum Mean Discrepancy} (MMD), and qualifying similarities of two time periods.
\textit{Maximum Mean Discrepancy} (MMD) is a distance metric between two probability distributions and the popular choice for dataset shift adaptation or domain adaptation~\cite{Yan-2017mind, wang2021-rethinking}.
Since the word sense change can be regarded as a problem of detecting shift, MMD is a suitable distance metric for comparing two time periods.
To my best knowledge, this is the first work using MMD for the task of word sense change detection.
Figure~\ref{fig:mainichi_heatmaps_MMD-Sense-Analysis} depicts the pairwise qualified similarities on multiple time periods.
Table~\ref{tab:word_sense_change_detection_result_ranking} represents selected words concerning the sense-aware dimension, and Figure~\ref{fig:word_sense_change_detection_sequence} shows the selected words' diachronic sense changes over multiple periods.

This paper is organized as follows. 
Section~\ref{sec:proposal-mmd-based-variable-selection} introduces the Maximum Mean Discrepancy (MMD) and the variable selection framework in optimizing an MMD estimator.
Section~\ref{sec:mmd-sense-analysis} presents the sense change detection algorithm, \texttt{MMD-Sense-Analysis}.
Section~\ref{sec:word_embedding_model} describes experimental results.

The codebase is available at the author's Github repository\footnote{\url{https://github.com/Kensuke-Mitsuzawa/mmd-tst-variable-selection-word-sense-change-analysis}}.


\section{Proposed Method}

The proposed algorithm \texttt{MMD-Sense-Analysis} selects variables of word embeddings that are involved with the sense changes (\textit{semantic aware dimensions} \cite{Aida-bollegala-2024-semantic}), and then the proposed algorithm selects vocabularies (words) significantly influenced by the selected variables.
In this paper, I call such variables \texttt{sense change variables}.

I, firstly, formalize preliminaries.
There is $T \in \mathbb{N}$ time periods to be studied, and word embedding vectors $E_{t} = \{e_1, ..., e_{V} \}, e \in \mathbb{R}^{D}$ are given for a time period index $t$, where $V$ is a set of vocabularies that the word embedding models.
Essentially, \texttt{MMD-Sense-Analysis} can handle any word embedding models, however, it is recommended to use a word embedding model designed for treating the time-space, such as \textit{PMI-SVD joint}~\cite{Aida-2021}.
\texttt{MMD-Sense-Analysis} selects \texttt{sense change variables} by the variable selection problem.
The variable selection problem is defined as selecting a subset $S \subset \{ 1, 2, \dots, D \}$ of variables out of the $D$ variables of word embedding vectors.
The selected variables $S$ represent indices of \texttt{sense change variables}.
For the variable selection, I use the approach of ``MMD Variable Selection''~\citep{mitsuzawa-2023}.

In the next section, I introduce Maximum Mean Discrepancy~\citep{Gretton-2012a} and the variable selection algorithm,
and then, \texttt{MMD-Sense-Analysis} is introduced in Section~\ref{sec:mmd-sense-analysis}.

\subsection{MMD Two-Sample Variable Selection}
\label{sec:proposal-mmd-based-variable-selection}

\textbf{Maximum Mean Discrepancy and MMD Variable Selection.}

Maximum Mean Discrepancy (MMD) is a kernel-based distance metric between probability distributions~\cite{Gretton-2012a} and a measure of distance between two probability distributions.
It relies on a positive semi-definite kernel~\cite{Hofmann-2008, Smola-2009} to map the distributions into a new space, a \textit{Reproducing Kernel Hilbert Space}~(RKHS), where probability distributions are represented as data points~\cite{Muandet-2017}.
A large MMD value indicates significant differences between two distributions.

Given i.i.d.~samples ${\bf X} = \{ \bx_1, \dots, \bx_n \} \stackrel{i.i.d.}{\sim} P$ and ${\bf Y} = \{\by_1, \dots, \by_n\} \stackrel{i.i.d.}{\sim} Q$, the MMD can be consistently estimated as  
\begin{equation} 
  \label{eq:mmd-unbiased-est}
  \begin{split}
    & \widehat{\rm MMD}^{2}_n( {\bf X}, {\bf Y}) := \frac{1}{n (n-1)} \sum_{i \not= i'} k(\bx_i, \bx_{i'})  + \frac{1}{n(n-1)} \sum_{ j \not= j'} k(\by_j, \by_{j'})  - \frac{2}{n^2} \sum_{i,j} k(\bx_i, \by_j).
  \end{split}
\end{equation}

Recently, two-sample variable algorithms based on MMD have been studied in the literature \citep{sutherland2017,wang2023variable,mitsuzawa-2023}.
Here, I explain the approach of \citep{mitsuzawa-2023}.

First, I introduce {\em Automatic Relevance Detection (ARD)} kernel as the kernel $k$ in Eq.~\eqref{eq:mmd-unbiased-est}, which is defined as 
\begin{equation} 
    \label{eq:ARD-kernel}
    k(\bx, \by) = \exp \left(- \frac{1}{D} \sum_{d=1}^{D} \frac{a_d^2 (x_d - y_d)^2}{\gamma^2_d} \right), \quad \bx:=(x_1,\threeDots,x_D)^{\intercal} \in \mathbb{R}^{D},\ \by:=(y_1,\threeDots,y_D)^{\intercal} \in \mathbb{R}^{D}
\end{equation}
where  $a_1,\threeDots,a_D \geq 0$ are called {\em ARD weights}, and $\gamma_1, \dots, \gamma_D > 0$ are constants to unit-normalize each variable. 
I compute $\gamma_1, \dots, \gamma_D$ by using the {\em dimension-wise median (or mean) heuristic}; see \cite[Appendix C]{mitsuzawa-2023} for details.

\cite{sutherland2017} optimizes the ARD weights $a_1, \dots, a_D$ in Eq.\eqref{eq:ARD-kernel} by maximizing the {\em test power} (i.e., the rejection probability of the null hypothesis $P = Q$ when $P \not=Q$) of a two-sample test based on the MMD estimate~Eq.\eqref{eq:mmd-unbiased-est}.
\cite{mitsuzawa-2023} proposes the variable selection by introducing $L_1$ regularization~\cite{Tibshirani-1996} to encourage sparsity for variable selection: 

\begin{equation}
  \label{eq:mmd-optimisation-problem}
  \underset{a \in \mathbb{R}^{\Dim}}{\min} -\log(\ell(a_1, \threeDots, a_{\Dim})) + \lambda \sum_{d=1}^{\Dim} |a_d|, \quad \text{where}\quad \ell(a_1, \threeDots, a_{\Dim}) := \frac{ \widehat{\rm MMD}^{2}_n( {\bf X}, {\bf Y})}{ \sqrt{\mathbb{V}_n({\bf X}, {\bf Y}) + C}}.
\end{equation}
In the above display, $\lambda > 0$ is a regularization constant, $C = 10^{-8}$ is a constant for numerical stability, and $\mathbb{V}_n({\bf X}, {\bf Y})$ is an empirical estimate of the variance of $\widehat{\rm MMD}^{2}_n( {\bf X}, {\bf Y})$; see \citep{mitsuzawa-2023} for the concrete form. 
The weights $a_1, \dots, a_D$ with larger $\ell(a_1, \dots, a_D)$ make the test power higher, thus making the two data sets ${\bf X}$ and ${\bf Y}$ more distinguishable. 
Minimizing the negative logarithm of $\ell(a_1, \dots, a_D)$ plus the $L_1$ regularization term $\lambda \sum_{d=1}^{\Dim} |a_d|$ thus yield {\em sparse} weights $a_1, \dots, a_D$ that distinguish ${\bf X}$ and ${\bf Y}$. 
Such weights can be used for selecting variables where the two data sets differ significantly. 

The regularization constant $\lambda$ controls the sparsity of optimized ARD weights, and an appropriate value of $\lambda$ is unknown beforehand, as it depends on the number of unknown active variables $S \subset \{1, \dots, D \}$ on which $P$ and $Q$ are different. 
\cite{mitsuzawa-2023} proposes a variable selection algorithm~\cite[Algorithm 2]{mitsuzawa-2023}, ``MMD Cross-Validation Aggregation'', selecting variables by aggregating variable selection results of various $\lambda$ values instead of specifying a $\lambda$.
I denote $\mmdCvAgg$.

\subsection{Detecting and Analyzing Word Sense Changes}
\label{sec:mmd-sense-analysis}

\textbf{Variable Selection.} 
I denote $S_{t, t'} = \mmdCvAgg$ as the procedure of selecting the \texttt{sense change variables} for the set of possible combination of time periods $\{ t, t' \in \{1, ..., T\} | t \neq t' \}$ of which size if $C^{T}_2$.
And I obtain $\setSelectedSenseChangeVariables$ that is a set of \texttt{sense change variables} for all possible combinations of time periods.

\textbf{Permutation Test.} \texttt{MMD-Sense-Analysis} conducts a permutation test for inspecting how the \texttt{sense change variables} $S_{t, t'}$ distinguish two dataset $E_t, E_{t'}$. 
The permutation test calculates the p-value.
The p-value represents the degree to which the observation data pair $E_t, E_{t'}$ is significantly against the null hypothesis, where the null-hypothesis is that word embedding distributions of two time periods are the same.
Intuitively, a lower p-value can indicate that the word sense between two time periods is different. 


\textbf{Global-Time Word Scoring.} 
When the variable selection is done for time period combinations, the scoring calculates a score representing the degree to which a vocabulary (word) $v$ is influenced by \texttt{sense change variables}. 
Given $\setSelectedSenseChangeVariables$, the scoring function is defined as
\begin{equation}
    \label{eq:word_sense_change_detection_score}
    \text{score}^{v}_{t} = \frac{1}{T^{\prime}} \sum_{\substack{t^{\prime} = 1 \\ t \neq t^{\prime}}}^{T^{\prime}} \text{cosine}(e_{v, t}^{S_{t, t^{\prime}}}, e_{v, t^{\prime}}^{S_{t, t^{\prime}}}),
\end{equation}
where $T^{\prime} = T - 1$ is the number of time periods without $t$, $t^{\prime} \in \{1, \threeDots, T^{\prime} \}$.
$e_{v, t}^{S_{t, t^{\prime}}}$ denotes a vector that selects \texttt{sense change variables} $S_{t, t'}$ from a word embedding vector $e$ at time $t$ for a vocabulary $v$.
The score range is $[0.0, 2.0]$.

This score expresses the average of cosine distances between the word embedding vectors using only \texttt{sense change variables}; therefore, a higher score value implies that a vocabulary $v$ is significantly related to the \texttt{sense change variables} at the time periods $t$.

\section{Demonstration of Word Sense Change Detection}
\label{sec:word_embedding_model}

I use the word embedding model \textit{PMI-SVD joint}~\cite{Aida-2021}.
The model is based on the PMI-SVD model~\citep{Levy-2014} that shows one can obtain the equivalent word embedding of skip-gram negative sampling~\citep{Mikolov-2013} by SVD matrix factorisation of the Pointwise Mutual Information (PMI) matrix.
The PMI-SVD joint model has a major difference from the original PMI-SVD model is that the PMI-SVD joint model can consider the sense change of words.
For the model construction, I follow the suggestion of \citep{kiyama-2025}.

\subsection{Japanese News Text Collection}


The dataset is a news text collection \textit{Mainichi Shimbun dataset}\footnote{\url{https://mainichi.jp/contents/edu/03.html}} written in Japanese by \textit{Mainichi Shinbun}\footnote{\jptext{毎日新聞社} in Japanese} news agency between 2003 and 2020.
A text collection during one year is treated as one period; the dataset is consisted of $18$ time periods.

Before training the word embedding model by \textit{PMI-SVD joint}~\cite{Aida-2021}, the tokenization is made.
Since the Japanese writing system does not have word dividers of white spaces, the text collection is tokenized by \textit{MeCab} tokenizer\footnote{\url{https://taku910.github.io/mecab/}} with a dictionary \textit{unidic}\footnote{\url{https://clrd.ninjal.ac.jp/unidic/en/}}.
In the preprocessing, sentences are filtered out if a sentence contains fewer than $20$ tokens.
Finally, the vocabulary set is constructed by tokens of which frequency is more than $100$ in text collections per year.
The constructed vocabulary set consists of $7,228$ vocabularies.
The word embedding model converts a word into a vector of $D=100$.

\textbf{Configuration of \texttt{MMD-Sense-Analysis}.} \texttt{MMD-Sense-Analysis} is applied on $153 = C^{18}_2$ pairs.

Filtering by \textit{Part-of-Speed}~(POS) and limiting the number of vocabularies is set.
This is because of MMD's computation cost $\mathcal{O}(n^2)$ where $n$ is the number of samples; the number of vocabularies in this context.
Nouns are chosen since nouns are more likely to convey the sense of the word, and they would sound proper to our research interests. 
After choosing the noun word, I randomly select $2,300$ vocabulary.

The $2,300$ vocabularies are split into training and test sets: $2,000$ vocabularies for selecting \texttt{sense change variables} and the rest $300$ for the permutation test.

\begin{figure}[!h]
    \centering
    \begin{minipage}{0.45\textwidth}
        \centering
        \includegraphics[width=\textwidth]{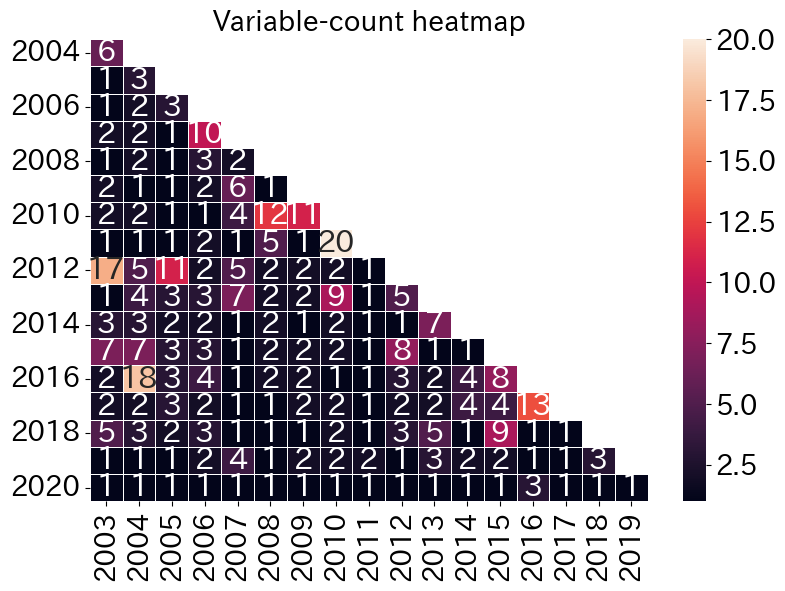}
        \label{fig:mainichi_word_sense_change_detection_result_variable_count}        
    \end{minipage}\hfill
    \begin{minipage}{0.45\textwidth}
        \includegraphics[width=\textwidth]{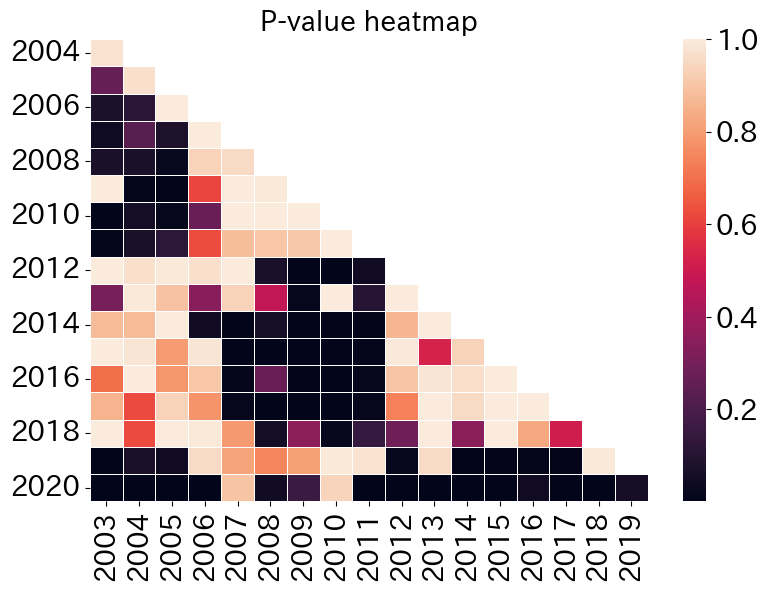}
        \label{fig:mainichi_word_sense_change_detection_result_pvalues}
    \end{minipage}\hfill
    \caption{
        (Left) Heatmap of \texttt{sense change variables}. The numbers in cells represent the number of selected \texttt{sense change variables}.
        (Right) Heatmap of p-values computed by the permutation test using the \texttt{sense change variables}.
        The horizontal and vertical labels show the period (year), and cells are pairwise comparisons of two time periods.
        For example, the right heatmap indicates that the word sense of the period ``2020'' is likely to be different from almost all other periods. 
        \label{fig:mainichi_heatmaps_MMD-Sense-Analysis}
    }
\end{figure}

\textbf{Results of Sense Change Variables.}
Figure~\ref{fig:mainichi_heatmaps_MMD-Sense-Analysis} represents heatmaps of counting selected \texttt{sense change variables} (left) on $153 = C^{18}_2$ pairs and p-values~(right).

\begin{figure}[!h]
    \centering    
    \begin{minipage}{0.45\textwidth}
        \centering
        \includegraphics[width=\textwidth]{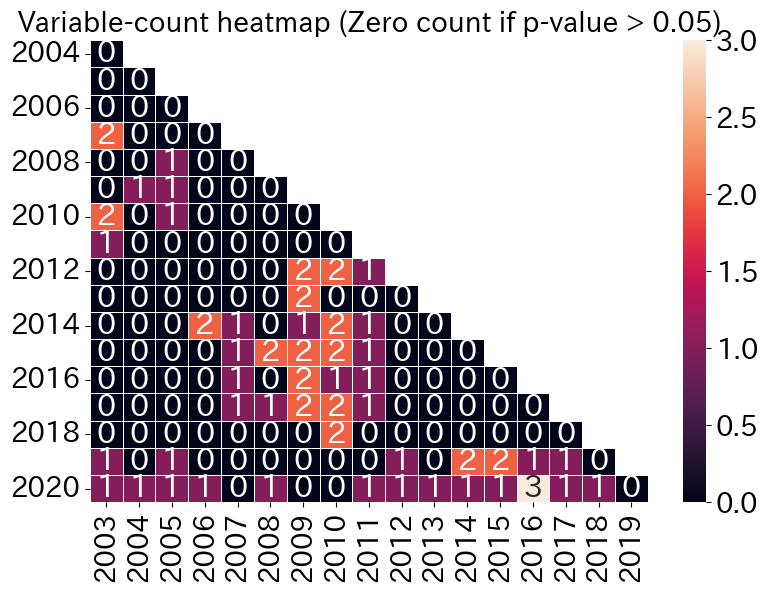}
    \end{minipage}\hfill
    \begin{minipage}{0.45\textwidth}
        \includegraphics[width=\textwidth]{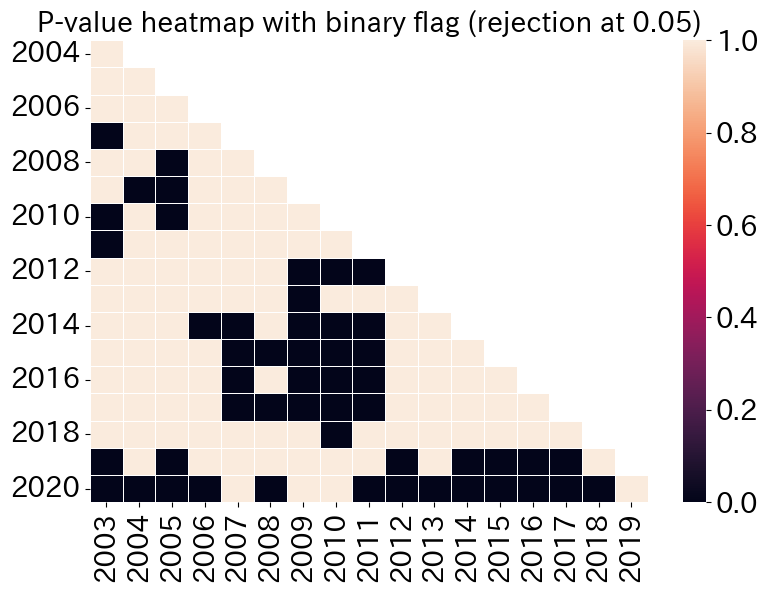}
    \end{minipage}\hfill

    \caption{
        Heatmaps of the selected \texttt{sense change variables} and p-values.
        The heatmaps show values at cells where p-value $<0.05$ of Figure~\ref{fig:mainichi_heatmaps_MMD-Sense-Analysis}(right).
        \label{fig:mainichi_heatmaps_MMD-Sense-Analysis-p-value-filter}
    }
\end{figure}

The heatmaps depict the following two tendencies: 1. the period 2020 frequently differs from other periods, 2. the period between 2009 and 2011 is often different from periods of 2012 and 2017.
The frequent dissimilarities of the period 2020 are considered to be due to COVID-19.
The period of 2020 is the year that COVID-19 pandemic started, and the Japanese government announced the state of emergency and lockdown on 7th April, 2020.
Thus, the new text collection in 2020 and the trained word embedding vectors can be significantly different from other periods.

The particularity of the time periods between 2009-2011 would be accounted for by two social backgrounds.
The first social background is the global depression caused by the bankruptcy of \textit{Lehman Brothers} announced on 15th September 2008, so-called the \textit{Lehman shock}~\cite{Wiggins-2019lehman}.
There would have been some time gap until the depression started to influence Japanese society~\cite{Shirota2020-performance}.
Since the word embedding model is trained by the news articles per year, the word embedding model of 2008 may not have learned the word sense about this depression, and the depression influenced the news text in 2009.

The second social background is \textit{the Great East Japan Earthquake} in March 2011.
The megathrust earthquake with the energy magnitude scales around $9.0$ caused the devastating tsunami disaster; Around $20,000$ casualties are reported~\cite{Mimura-2011damage} by this disaster, and the tsunami triggered an incident of the \textit{Fukushima Daiichi nuclear disaster}.
This natural disaster and the following incident caused huge economic loss~\cite{Kajitani-2014estimation} and enormous social impact~\cite{Okuyama2017-influence}.
Considering these social impacts, it is natural that the disaster is the major topic in the text collection of 2011, and the word embedding vectors are significantly different from other periods.

\textbf{Results of Global-Time Word Scoring.} Based on the results of \texttt{sense change variables}, I focus on the periods of 2009, 2010, 2011, and 2020.
Table~\ref{tab:word_sense_change_detection_result_ranking} represents the top three words with high scores of \texttt{Global-Time Word Scoring} in Eq.\eqref{eq:word_sense_change_detection_score} for the these years.

\begin{table}[!ht]
    \centering
    \caption{
        The top three words with high scores in the sense change detection task for the years 2009, 2010, 2011, and 2020.
        The extracted Japanese words are translated into English, and the original word is in the square brackets,
        and the values in the parentheses are the global-time word score computed by Eq.\eqref{eq:word_sense_change_detection_score}.
        \label{tab:word_sense_change_detection_result_ranking}
    }
    \begin{tabular}{l|lll}
    \toprule
        year /\ rank & 1 & 2 & 3 \\
        \midrule
        2009 & economic depression [\jptext{不況}] (0.872) & golf [\jptext{ゴルフ}] (0.803) & carbon [\jptext{炭素}] (0.801) \\ 
        2010 & economic depression [\jptext{不況}] (0.596) & fish boat [\jptext{漁船}] (0.572) & life [\jptext{住}] (0.569) \\
        2011 & disaster victim [\jptext{被災}] (2.0) & evacuation [\jptext{避難}] (2.0) & recovery [\jptext{復旧}] (2.0) \\
        2020 & health condition [\jptext{具合}] (1.988) & handling [\jptext{対処}] (1.988) & hospital care [\jptext{入院}] (1.984) \\

        \bottomrule
    \end{tabular}
\end{table}

These highlighted words may describe the social backgrounds of the target years.
As discussed above, \texttt{MMD-Sense-Analysis} sounds to detect variables in the periods of 2009 and 2010 due to the Lehman shock; The period of 2011 is due to the Great East Japan Earthquake, and the period of 2020 is due to the COVID-19 pandemic.
The listed words are related to these events, thus these words explain the selected \texttt{sense change variables}.

In the ranking of 2009, there are words \textit{golf} and \textit{carbon}.
No clear and certain explanations can be provided for these words since I do not have access to the text collection dataset for training the word embedding model.
Yet, the keyword \textit{golf} may account for by the youngest professional golfer champion, Ryo Ishikawa, who won the Japan Golf Tour in 2009, and the keyword \textit{carbon} may be related to policies about the global warming issue and the news that the Japanese government achieved the target reduction value in 2009.

\begin{figure}
    \centering
    \begin{minipage}{0.45\textwidth}
        \centering
        \includegraphics[width=\textwidth]{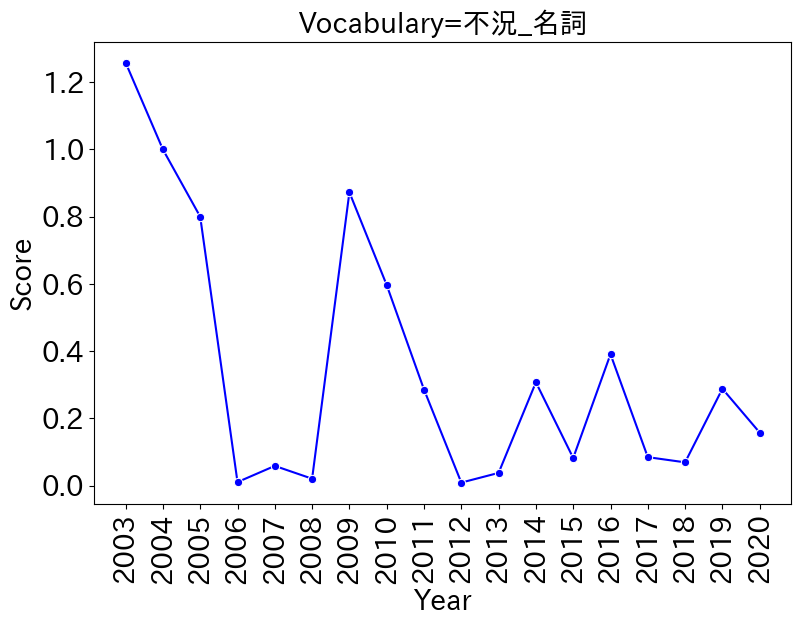}
        \subcaption{\textit{economic depression} \jptext{(不況)}}
    \end{minipage}\hfill
    \begin{minipage}{0.45\textwidth}
        \centering
        \includegraphics[width=\textwidth]{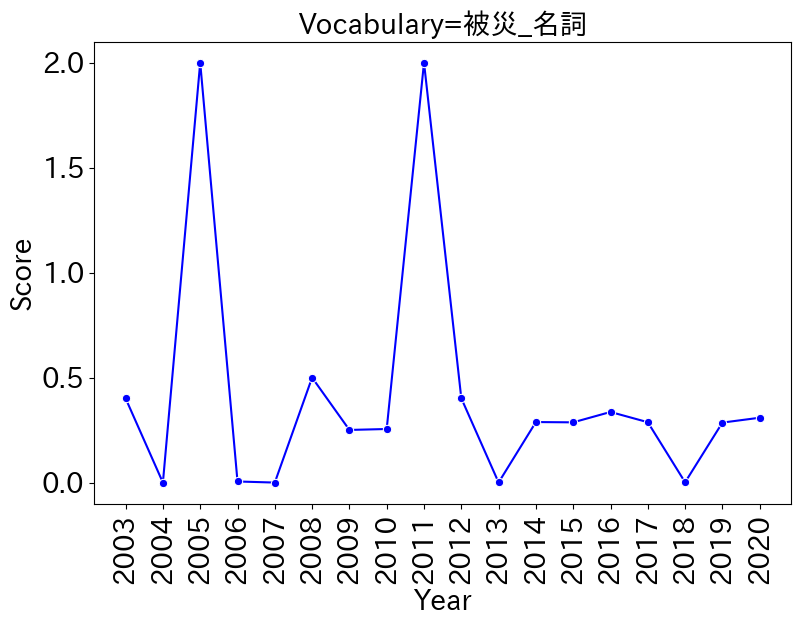}
        \subcaption{\textit{disaster victim} \jptext{(被災)}}
    \end{minipage}



    \caption{
        The diachronic word sense analysis using the \texttt{sense change variables}.
        The horizontal axis is the year label.
        The vertical axis is the global-time word score computed by Eq.\eqref{eq:word_sense_change_detection_score}.
        A higher value indicates that the word sense of a period (year) is particularly different from other periods.
        }
    \label{fig:word_sense_change_detection_sequence}
\end{figure}

\textbf{Word Sense Change Analysis.} The \texttt{global-time word score} in Eq.\eqref{eq:word_sense_change_detection_score} computes a score at each period $t$ for a vocabulary $v$.
To study further the sense changes of selected keywords, I plot the sequences of \texttt{global-time word score} between 2003 and 2020 in Figure~\ref{fig:word_sense_change_detection_sequence}.
The keyword \textit{economic depression} (Figure~\ref{fig:word_sense_change_detection_sequence}, left) has a high word score in 2009, as discussed in Table~\ref{tab:word_sense_change_detection_result_ranking}.
The peak of 2003 is considered to be involved in the long-continued economic depression in Japan since the economic outbreak in 1991, which is called \textit{Japan's Lost Decade}~\citep{Callen-2003}; The year of 2003 is the first year that the long economic depression ended and economic growth restarted again.


The peak of \textit{disaster victim} (Figure~\ref{fig:word_sense_change_detection_sequence}, right) in 2011 is due to the Great East Japan Earthquake.
This high score in 2005 would be accounted for by multiple incidents, not only in Japan but also all over the world.
Table~\ref{tab:table-event-victim-2005} lists major incidents, terrorism attacks, and natural disasters.
Considering these frequent and various types of disastrous events in 2005, there would be various contexts that make co-occurrences with \textit{disaster victim}, and the embedding model would learn differently from other periods.

\begin{table}[!h]
    \centering
    \caption{
        List of incidents and natural disasters that occurred in 2005.
        \label{tab:table-event-victim-2005}
    }
    \begin{tabularx}{\textwidth}{llX}
    \toprule
    Incident & Date & Summary \\
    \midrule
    Valentine's Day bombings & February 2005 & Terrorism attack in multiple cities in Philippine. \\    
    Amagasaki derailment & April 2005 & Train derailment accident causing mass casualties. \\
    London bombings & July 2005 & Terrorism attack in the central London. \\    
    Hurricane Katrina & August 2005 & Powerful, devastating and historic tropical cyclone in the city of New Orleans. \\
    Kashmir earthquake & October 2005 & Earthquake causing massive damages in wide area around the Kashmir region. \\
    \bottomrule
    \end{tabularx}
\end{table}


\subsection{Clean Corpus of Historical American English (CCOHA)}
\label{sec:result-cchoha}

\textit{CCOHA}~\citep{Alatrash-2020} is a large scale corpus containing a collection of American English texts from 1810 to 2010.
For the word embedding model, I segment the corpus into 10-year time periods.
I did not use the time periods of 1820s, following the suggestion~\cite{kiyama-2025} and therefore, there are 19 time periods. 
The target words are those that have more than 100 frequencies in each period, and 505 words are selected.
The word embedding model is trained by the PMI-SVD join~\cite{Aida-2021}, and the dimension of vectors is $D=100$.

For \texttt{MMD-Sense-Analysis}, 400 words are used for the variable selection, and 105 words for the permutation test.


\textbf{Results of Sense Change Variables.}

\begin{figure}[!h]
    \centering
    \begin{minipage}{0.45\textwidth}
        \centering
        \includegraphics[width=\textwidth]{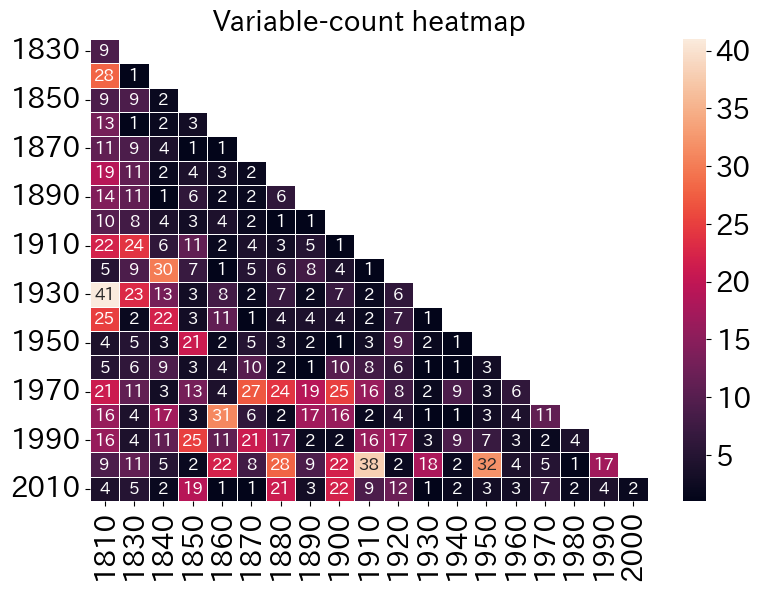}
        \label{fig:ccoha_word_sense_change_detection_result_variable_count}
    \end{minipage}\hfill
    \begin{minipage}{0.45\textwidth}
        \includegraphics[width=\textwidth]{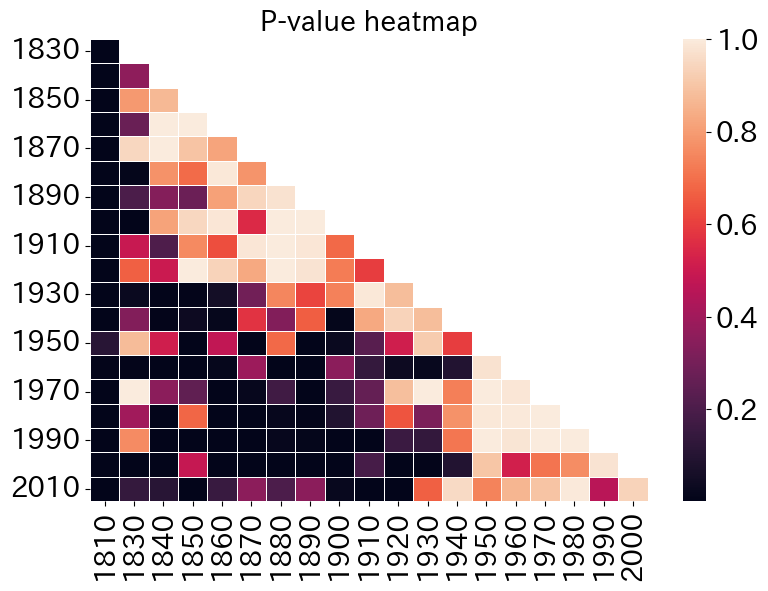}
        \label{fig:ccoha_word_sense_change_detection_result_pvalues}
    \end{minipage}\hfill


    \begin{minipage}{0.45\textwidth}
        \centering
        \includegraphics[width=\textwidth]{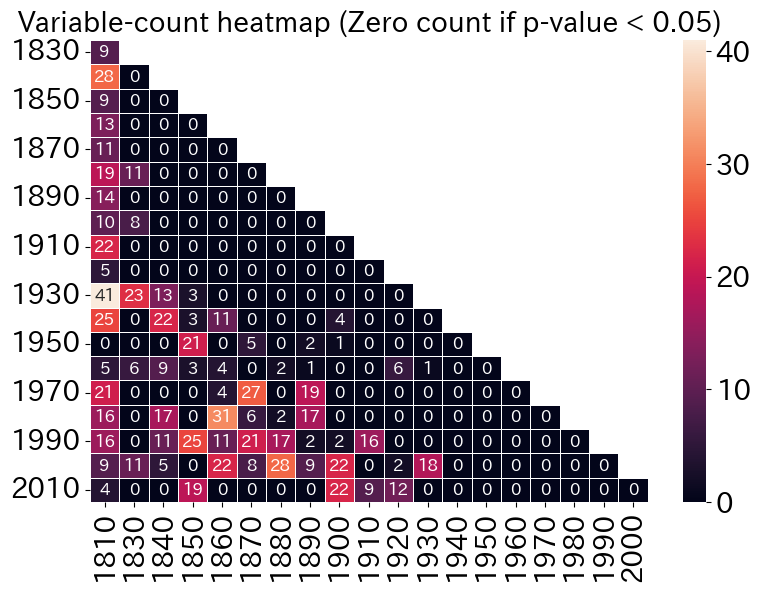}
    \end{minipage}\hfill
    \begin{minipage}{0.45\textwidth}
        \includegraphics[width=\textwidth]{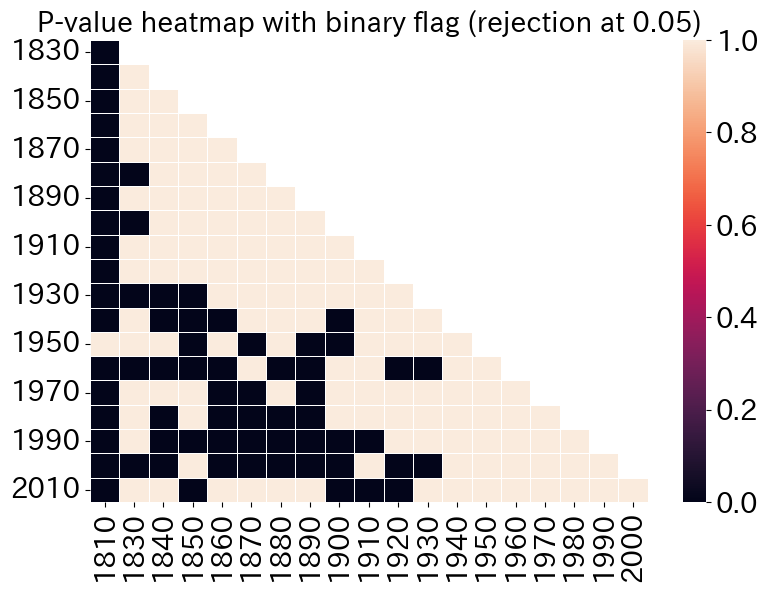}
    \end{minipage}\hfill
    \caption{
        The result of the sense change detection task.
        \label{fig:coha_heatmaps_MMD-Sense-Analysis}
    }
\end{figure}

Figure~\ref{fig:coha_heatmaps_MMD-Sense-Analysis} represents heatmaps of selected \texttt{sense change variables} and p-values. 
In the heatmap, there is a sub-block that holds low p-values and more selected variables at the horizontal axis of 1860-1890 and the vertical-axis of 1950-2000.
\cite{Bochkarev-2014universals} reports that the English lexicon has experienced a high rate of variability since the 1850s.
The selected \texttt{sense change variables} may explain the high variability.

\textbf{Results of Global-Time Word Scoring.}
Table~\ref{tab:word_sense_change_detection_result_ranking} represents the top three words with high scores of \texttt{Global-Time Word Scoring} in Eq.\eqref{eq:word_sense_change_detection_score} for the following periods: 1850s, 1860s, 1980s, and 1990s.

\begin{table}[!ht]
    \centering
    \caption{
        \texttt{Global-Time Word Scoring} results based on the selected \texttt{sense change variables}.
        \label{tab:ccoha_word_sense_change_detection_result_ranking}
    }
    \begin{tabular}{l|lll}
    \toprule
        year /\ rank & 1 & 2 & 3 \\
        \midrule
        1850 & representative (1.166) & political (1.094) & country (1.023) \\
        1860 & declare (1.152) & president (1.092) & observe (1.039) \\
        \hline
        1980 & concern (1.009) & attempt (0.963) & regard (0.947) \\
        1990 & seek (0.989) & necessity (0.954) & suppose (0.922) \\
        \bottomrule
    \end{tabular}
\end{table}

The 1850s and 1860s are eras of the \textit{American Civil War} (April 1861 - May 1865).
Considering this social background, the selected words such as ``political'', ``country'', ``declare'', ``president'' sound to have a particular sense related to the civil war.

On the other hand, the selected words in the 1980s and 1990s are insufficient to explain the social background.
These selected words can be verbs, and it is hard to interpret the sense without related nouns.
The insufficient explanation can be due to the small vocabulary size in this experiment.
Since the prepared vocabulary size is only $505$ words, it is really challenging to explain the \texttt{sense change variables} and infer the sense changes of words.
This result empirically tells me that \texttt{MMD-Sense-Analysis} requires abundant vocabularies.

\textbf{Word Sense Change Analysis.}

\begin{figure}[h]
    \centering
    \includegraphics[scale=0.5]{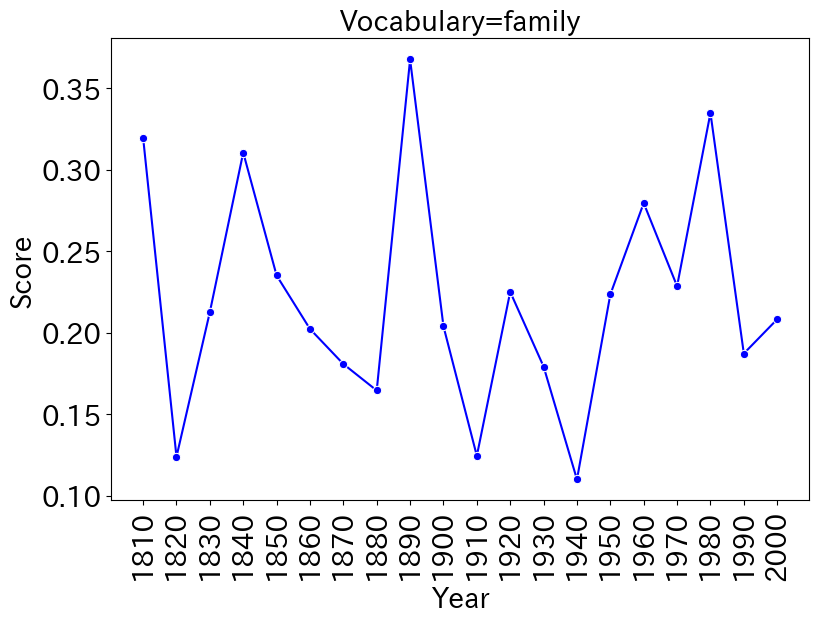}
    \caption{
        The diachronic word sense analysis using the \texttt{sense change variables}.
        The horizontal axis is the year label.
        The vertical axis is the global-time word score computed by Eq.\eqref{eq:word_sense_change_detection_score}.
        A higher value represents a particular word sense at a year.
        \label{fig:coha_word_analysis_keyword_family}
    }
\end{figure}

The paper~\cite{Poole-2023lexical} conducts diachronic variation analysis on American English, and points out that the keyword ``family'' has high variance in the 1980s.
Figure~\ref{fig:coha_word_analysis_keyword_family} shows diachronic sequences of the score at Eq.~\eqref{fig:coha_word_analysis_keyword_family}, and representing the high score in the 1980s.

\section{Conclusion}
\label{sec:conclusion}

In this paper, I propose \texttt{MMD-Sense-Analysis}: an algorithm for sense change detection.
\texttt{MMD-Sense-Analysis} is supported by the variable selection approach using Maximum Mean Discrepancy (MMD) that is suitable for dataset adaptation or domain adaptation.
By the variable selection, \texttt{MMD-Sense-Analysis} can select the semantic aware variables that are significantly involved in the sense changes and qualify the similarities of multiple time periods.
Based on the selected semantic aware variables, \texttt{MMD-Sense-Analysis} selects words concerning the semantic aware variables.

The empirical results show that the similarities of time periods can explain the social backgrounds, and high-scoring words can clearly describe the social backgrounds as well as the qualified similarities.
Yet, the empirical result shows that it is challenging to explain the social backgrounds when the available vocabulary is insufficient.

As future work, I plan to evaluate \texttt{MMD-Sense-Analysis} on a benchmark dataset for the sense change detection.

\paragraph{Acknowledgements}

This work has been supported by the French government, through the NIMML project (ANR-21-CE23-0005-01).
I gratefully acknowledge Mr. Hajime Kiyama of Hitotsubashi University for generously providing the word embedding models used in this study.

\bibliographystyle{plain}
\bibliography{bibliography}


\end{document}